\pdfoutput=1

\documentclass[11pt]{article}

\usepackage[preprint]{acl}

\usepackage{times}
\usepackage{latexsym}

\usepackage[T1]{fontenc}

\usepackage[utf8]{inputenc}

\usepackage{microtype}

\usepackage{subcaption}
\usepackage{cleveref}
\usepackage{multirow}
\usepackage{multicol}
\usepackage{booktabs}
\usepackage{makecell}

\crefname{section}{Section}{Sections}
\crefname{table}{Table}{}
\crefname{figure}{Figure}{}
\crefname{section}{\S}{\S\S}
\Crefname{section}{\S}{\S\S}
\crefname{appendix}{Appendix}{Appendices}
\Crefname{Appendix}{Appendix}{}

\usepackage{inconsolata}

\usepackage{graphicx}

\title{Lost in Inference: Rediscovering the Role of Natural Language Inference for Large Language Models}

\author{Lovish Madaan$^{\infty}$$^{\diamondsuit}$ \quad
David Esiobu$^{\infty}$ \quad
Pontus Stenetorp$^{\diamondsuit}$$^{\spadesuit}$ \quad \\
\textbf{Barbara Plank}$^{\clubsuit}$$^{\varbigtriangleup}$ \quad
\textbf{Dieuwke Hupkes}$^{\infty}$ \\
\vspace{1mm}
$^{\infty}$Meta \quad $^{\diamondsuit}$University College London \\
$^{\spadesuit}$Research and Development Center for LLMs, National Institute of Informatics \\
$^{\clubsuit}$MaiNLP, Center for Information and Language Processing, LMU Munich, Germany\\
$^{\varbigtriangleup}$Munich Center for Machine Learning (MCML), Munich, Germany\\
\texttt{\{lovish,dieuwkehupkes\}@meta.com}
}

\begin{document}
\maketitle

\begin{abstract}
In the recent past, a popular way of evaluating natural language understanding (NLU), was to consider a model's ability to perform natural language inference (NLI) tasks. 
In this paper, we investigate if NLI tasks, that are rarely used for LLM evaluation, can still be informative for evaluating LLMs.
Focusing on five different NLI benchmarks across six models of different scales, we investigate if they are able to discriminate models of different size and quality and how their accuracies develop during training.
Furthermore, we investigate the extent to which the softmax distributions of models align with human distributions in cases where statements are ambiguous or vague.
Overall, our results paint a positive picture for the NLI tasks: we find that they are able to discriminate well between models at various stages of training, yet are not (all) saturated.
Furthermore, we find that while the similarity of model distributions with human label distributions increases with scale, it is still much higher than the similarity between two populations of humans, making it a potentially interesting statistic to consider.
\end{abstract}

\section{Introduction}

Before the state-of-the-art (SoTA) in NLP was constituted almost exclusively by large language models (LLMs), a popular way of evaluating models' understanding of natural language was to consider their ability to perform \emph{natural language inference} (NLI) tasks \citep[most famously,][]{bowman-etal-2015-large,williams-etal-2018-broad}.
Motivated by the idea that concepts such as entailment and contradiction are central to many aspects of language meaning \citep{bowman-etal-2015-large}, in NLI tasks, a model is asked to judge the relationship between the meaning of two sentences, typically chosing between entailment, contradiction, and no relationship.
Included in the then widely-used natural language understanding benchmark GLUE \citep{wang2019glue}, the NLI benchmark \emph{Multi-Genre Natural Language Inference}  \citep[MNLI,][]{williams-etal-2018-broad} was up until relatively recently one of the most popular benchmarks to evaluate language models, and is -- with over 600 citations to date in 2024 -- well-cited even in the recent past.

However, with the arrival of LLMs, MNLI and other datasets have lost their spot on the SoTA leaderboards.
With the exception of \citet{brown2020language},  who reported low scores for GPT-3 on the NLI benchmark \emph{Adversarial NLI} \citep[ANLI,][]{nie-etal-2020-adversarial}, not a single LLM release paper considers an NLI benchmark in their evaluation suite.\footnote{Some recent papers do report low scores for LLMs, for e.g.\ GPT3.5 on XNLI \citep{ohmer2024form,ohmer-etal-2023-separating} and Llama 1 models on ANLI, HANS, and MNLI \citep{mccoy-etal-2019-right,weber-etal-2023-mind}.}
In this paper, we investigate why.
Are NLI benchmarks simply not suitable to evaluate modern-day LLMs? 
Are their examples too difficult or too easy?
Are their scores not informative?
Or do they, in fact, still provide a useful signal?

Focusing on five different NLI benchmarks across six different models, we first show that they provide signal and are able to discriminate models of different size and quality (\cref{subsec:fully_trained}), but that one or more few-shot examples are needed to obtain a somewhat reasonable accuracy for models of any size.
We furthermore show that performance on the datasets develop steadily during training, albeit with some fluctuations, making the benchmarks suitable for monitoring training progress (\cref{subsec:during_training}).
For some of the benchmarks, accuracies of the best models are reaching 80-90\%, for ANLI \citep{nie-etal-2020-adversarial}, however, the accuracy of even the best model does not exceed 70\%.
Furthermore, we show that high scores are not caused by data contamination (\cref{subsec:contamination}).

Next, we investigate the extent to which improvement on the higher-scored benchmarks is still possible (\cref{subsec:saturation}).
In a manual analysis, we find that, for the best performing model, examples with `incorrect' predictions are rarely in fact incorrect; most concern questions on which humans may disagree as well.
Motivated by this result, we further explore this topic by considering the ChaosNLI benchmark \citep{nie-etal-2020-learn}, which contains 100 human annotations for over 4500 samples for three of the benchmarks we consider (\cref{subsec:chaosnli_dist}).
We find that accuracies (as computed on the majority label) are higher if the entropy of the human labels is low; when humans disagree, models are more likely to select one of the less preferred labels.
Lastly, we consider the distributional differences between model outputs and human labels, as measured by the Jensen Shannon Divergence (JSD) between the human label distributions and the models' softmax distributions over the possible answers.
We find that the JSDs are lower than the ones reported by \citet{nie-etal-2020-learn} for the previous generation of models, and they are also better than chance distributions. 
However, they are still substantially higher than the JSD between two populations of humans\footnote{For further evidence to this claim, see \citet{baan-etal-2022-stop}.}, bearing interesting implications on the viablity of using an ensemble of LLM judges as a `jury' \citep[e.g.][]{verga2024replacing}.
Interestingly, contrary to the findings of \citet{nie-etal-2020-learn}, we observe an effect of scale and model quality: JSD shows a clear decrease during training, and larger models have lower JSD than smaller models.

In sum, we find that NLI benchmarks are still useful for model development and improvement.
Specifically, they are able to discriminate between models of different scale and quality, develop steadily during training, and are not completely saturated.
Furthermore, as even the best models are still far away from human performance in this respect, we see promise in monitoring the development of the distributional differences between models and humans, both during and after training.

\section{Related work}

Before presenting our analysis, we first discuss NLI or \emph{recognising textual entailment} (RTE) tasks (\cref{related:nli}) and touch upon the related topic of subjectivity for NLP tasks (\cref{related:subjectivity}) before presenting our analysis.

\subsection{RTE tasks and their results}
\label{related:nli}

In RTE tasks -- also referred to as `natural language inference' (NLI) tasks, models are asked to judge whether the meaning of one sentence can be inferred from the meaning of another.
Because the concept of entailment (and contradiction) are considered central to many aspects of language meaning \citep[e.g.][]{bowman-etal-2015-large}, such tasks were for some period considered an important task to determine whether one model could understand language better than another \citep{poliak-2020-survey}.
Over the years, many different, increasingly more difficult NLI tasks have been proposed in the literature.
Included in the popular benchmarks GLUE \citep{wang2019glue} and SuperGLUE \citep{wang2019superglue}, the benchmarks MNLI \citep{williams-etal-2018-broad} and RTE \citep{dagan2005pascal} and RTE were used to claim SoTA by many influential models such as BERT \citep{devlin2019bert} and T5 \citep{raffel2023t5}.
When performance on MNLI and RTE started to saturate, several adversarial NLI benchmarks were introduced, such as ANLI \citep{raffel2023t5} and HANS \citep{mccoy-etal-2019-right}, on which BERT-style models performed poorly compared previous datasets.
For LLMs, NLI benchmarks are rarely used.
Of all big LLM releases, only GPT-3 \citep{brown2020language} reported an NLI score, and only on one partition of ANLI.
They concluded that NLI is a difficult task for general purpose LLMs.
Similar trends were observed by \citet{ohmer2024form} and \citet{weber-etal-2023-mind} for decoder-only LLMs on various NLI tasks, in part motivating the study presented here.

\subsection{Subjectivity in NLP tasks}
\label{related:subjectivity}
Another line of work relevant to ours considers the behaviour of models in cases where humans disagree on the correct label for a particular sample \citep[for an overview, see][]{plank-2022-problem}.
The ground truth labels for NLP benchmarks are often decided according to the majority label by human annotators. 
This simplifies the data annotation process, while also making the evaluation easier. 
However, several previous studies have noted that human disagreements in annotations for NLP datasets reflect the lack of a single ground truth label, rather than than noise in the annotation process \citep[e.g.][]{de-marneffe-etal-2012-happen, plank-etal-2014-learning,pavlick-kwiatkowski-2019-inherent,nie-etal-2020-learn}and that we could benefit from embracing rather than resolving it \citep{plank-2022-problem,aroyo2015truthIA}.

For NLI, label disagreements were captured in more detail in the dataset ChaosNLI \citep{nie-etal-2020-learn}, comprising of 100 annotations for each sample for a subset of three benchmarks -- MNLI, SNLI, and $\alpha$NLI.
As human disagreements are ubiquitous yet mostly ignored in LLM evaluation, in our study, we analyse not only the model accuracies, but also the relationship between these disagreements and models' probability distributions across the different labels.
A study with a similar aim is presented by \citet{chen2024seeingbig}. They explore whether the softmax probability distributions solicited by a few explanations from two LLMs (Mixtral and Llama 3) can approximate human judgement distributions (HJD) on the ChaosNLI and VariErr NLI \citep{weber-genzel-etal-2024-varierr} benchmarks.
Similarly, \citet{baan-etal-2024-interpreting} and \citet{lee-etal-2023-large} compare human and model judgement distributions on ChaosNLI, finding that models fail to capture human distributions in LLMs and model confidences. 
Furthermore, \citet{baan-etal-2024-interpreting} argue how the softmax probability distribution can be interpreted as both an approximation to human label distribution and confidence estimation in language models.

\section{Setup}

Before reporting our results, we first describe the benchmarks and models we consider in our experiments and describe our evaluation procedure.

\subsection{Benchmarks}
In our experiments, we consider five NLI benchmarks as briefly described below. 
A more elaborate description can be found in \cref{app:benchmarks}.

\vspace{-1mm}
\paragraph{SNLI}
The Stanford Natural Language Inference (SNLI) is one of the first large-scale NLI datasets for NLP evaluation, sourced via Amazon Mechanical Turk.
We use the dev set of the corpus, which comprises 10K examples.

\vspace{-1mm}
\paragraph{MNLI}
The Multi-Genre NLI (MNLI) corpus \citep{williams-etal-2018-broad} was introduced as an alternative to SNLI that captures more genres and challenging examples. For MNLI too we consider the 10K examples dev set of the corpus.

\vspace{-1mm}
\paragraph{HANS}
The adversarial dataset Heuristic Analysis for NLI Systems \citep[HANS,][]{mccoy-etal-2019-right} is generated to contain examples that cannot be solved through heuristics like lexical overlap, and contains 30K examples.

\vspace{-1mm}
\paragraph{ANLI}
The second adversarial dataset we consider is Adversarial NLI, or ANLI \citep{nie-etal-2020-adversarial}.
The dataset, created with the primary aim to make SoTA models fail, is sourced iteratively with a human-in-the-loop setup.
In our experiments, we consider the dev set of iteration 3, the most challenging set of the benchmark.

\vspace{-1mm}
\paragraph{$\alpha$NLI}
$\alpha$NLI or abductive NLI \citep{bhagavatula2020abductive} has a different setup than the previous benchmarks, focusing specifically on \emph{abductive reasoning}.
Each sample consist of a pair of observations at two consecutive times, a plausible hypothesis that explains tho two observations, and an implausible hypothesis that does not (or to a lesser extent).
The model has to select the most plausible hypothesis among the two choices.
The dataset contains 1500 examples.

\vspace{-1mm}
\paragraph{ChaosNLI}
Lastly, we consider ChaosNLI \citep{nie-etal-2020-learn}, which contains 100 additional (human) labels for 1500 examples for each of SNLI, MNLI, and $\alpha$NLI.

\subsection{Models} For each of these benchmarks, we compute and analyse scores for two different model families: Llama \citep{dubey2024llama} and Mistral \citep{jiang2023mistral, jiang2024mixtral}. 
Specifically, we use Meta-Llama 3.1 \{8, 70, 405\}B from the Llama series, and Mistral 7B / Mixtral 8x\{7, 22\}B from the Mistral series.
We limit our analysis to pre-trained base models and leave the analysis on post-trained / instruct models to future work.

\subsection{Evaluation details}
As nowadays common for pre-trained models, we consider the choice-based rather than generative evaluation setup for all the tasks \citep{dubey2024llama,mmluhfleaderboard}. 
In this setup, the model is presented with the few shot examples (if present) along with the question and the available choices in a multiple choice set-up, and is then asked to predict the correct letter choice.
Since there are only a limited number of choices depending on the task (two or three), we append the these choices to the prompt and compute the negative log likelihood (NLL) over the letter choice. 
We then choose the option which has the lowest NLL as the model's prediction. 
The prompt templates for all tasks are detailed in \cref{tab:prompt_template}. 
We use simplistic prompt templates for each of the tasks without much prompt tuning, exploiting the finding presented by \citet{dubey2024llama} that the Llama models are usually robust to prompt texts in the choice task evaluation setup.

\section{Results}

We now present our results, focusing in particular on whether NLI benchmarks provide a discriminative signal for fully trained out models (\cref{subsec:fully_trained}), how their performance develops during training (\cref{subsec:during_training}), and what is the impact of evaluation data contamination (\cref{subsec:contamination}). 
Next, we investigate whether there is still room for improvement (\cref{subsec:saturation}) and how model judgements compare to human judgements for ambiguous or vague questions (\cref{subsec:chaosnli_dist}).

\subsection{Informativeness for fully trained models}\label{subsec:fully_trained}

\begin{figure*}[t]
    \begin{subfigure}[b]{\textwidth}
        \hspace{5mm}
        \includegraphics[width=0.95\textwidth]{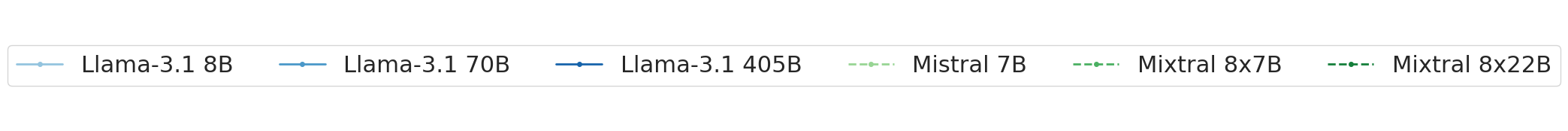}
        \vspace{-3mm}
    \end{subfigure}\\
    \begin{subfigure}[b]{0.217\textwidth}
    \centering
    \includegraphics[height=3.45cm]{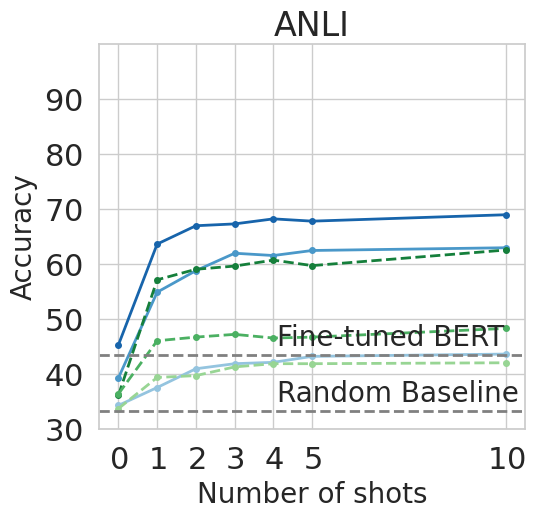}
    \caption{ANLI}
    \end{subfigure}
    \label{fig:anli}
    \begin{subfigure}[b]{0.19\textwidth}
    \centering
    \includegraphics[height=3.45cm, trim=25mm 0 0 0, clip]{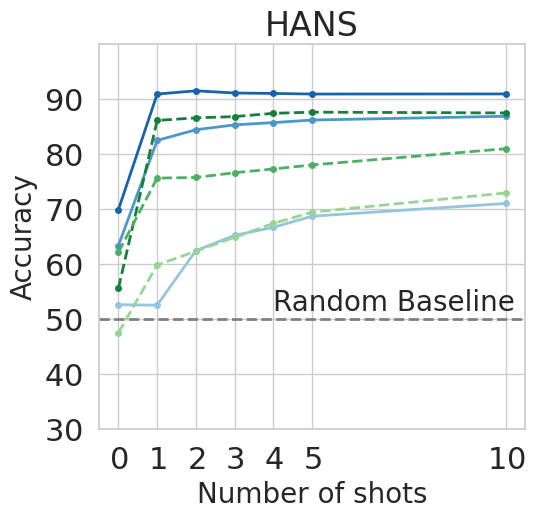}
    \caption{HANS}
    \label{fig:hans}
    \end{subfigure}
    \begin{subfigure}[b]{0.19\textwidth}
    \centering
    \includegraphics[height=3.45cm, trim=25mm 0 0 0, clip]{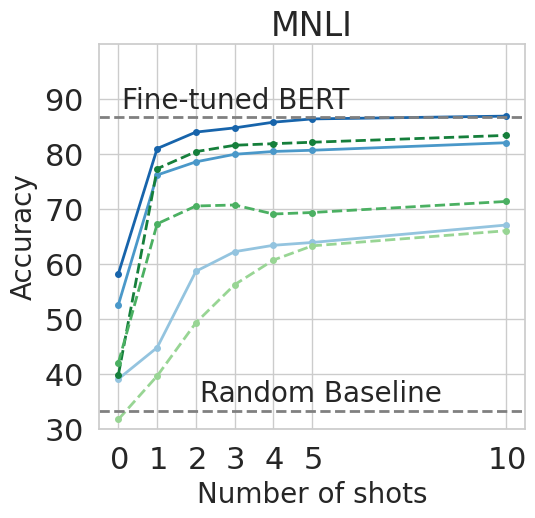}
    \caption{MNLI}
    \label{fig:mnli}
    \end{subfigure}
    \begin{subfigure}[b]{0.19\textwidth}
    \centering
    \includegraphics[height=3.45cm, trim=25mm 0 0 0, clip]{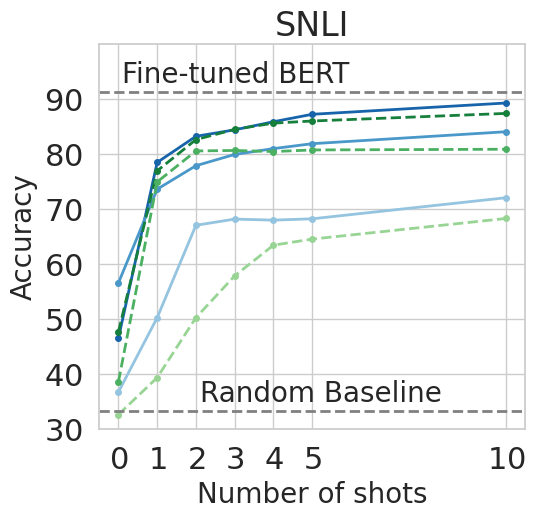}
    \caption{SNLI}
    \label{fig:snli}
    \end{subfigure}
    \begin{subfigure}[b]{0.19\textwidth}
    \centering
    \includegraphics[height=3.45cm, trim=25mm 0 0 0, clip]{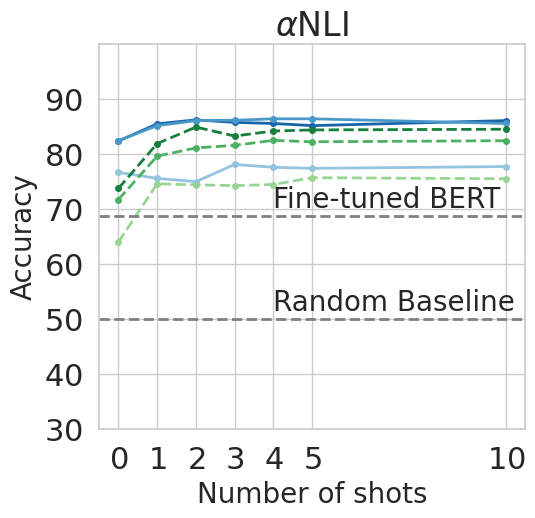}
    \caption{$\alpha$NLI}
    \label{fig:alphanli}
    \end{subfigure}
    \caption{\textbf{Performance across shots.} We show the accuracies for six fully pre-trained models on the five NLI benchmarks. Dashed lines indicate random and finetuned-BERT baselines.}\label{fig:shot_performance}
\end{figure*}

To get an overall estimate of the difficulty of NLI and the extent to which it can discriminate models, we first consider the performance of fully pre-trained models from the respective model series.

\paragraph{Accuracy across shots}
For each of the models, we compute results with a variable number of shots.\footnote{For each number of few-shot examples, samples were chosen randomly from the dev or train set once and kept fixed throughout the experiments.}
We report the results in \cref{fig:shot_performance}, along with a chance baseline and the results of a fine-tuned BERT model.
For all models, we observe rather poor zero-shot performance for all tasks, except $\alpha$NLI, confirming previously reported results by \citet{ohmer2024form}, \citet{weber-etal-2023-mind}, and \citet{dutt-etal-2024-investigating}.
When more few-shot examples are added, performance starkly improves.
Even with just one in-context example, the performance is significantly better than the zero shot accuracy.
Adding more than three or four examples marginally improves performance and saturates around ten few-shot examples. 
Among the five benchmarks, the most challenging benchmark is ANLI.
Although the larger models in the Llama and Mistral series far outperform the finetuned BERT baseline, they do not exceed 70\% accuracy -- to some extent confirming the difficulty of the benchmark reported by \citet{brown2020language}.

\paragraph{Model discriminability}
For discriminability, virtually all benchmarks provide a clear gap between the smaller and larger models for both families of models.
For example, for the Llama series, 405B performs the best followed by the 70B and then the 8B model.
Though these three models are trained on the same amount of text tokens, performance clearly improves with scale.
The exception to this pattern is $\alpha$NLI, which appears to be near-saturated already at 70B with an accuracy of around 85\%.
We conclude from these results, that the benchmarks provide a useful signal to compare trained-out models, though it is unclear to what extent their performance has saturated, which we discuss in more detail in \cref{subsec:saturation}.

\subsection{Informativeness during training}\label{subsec:during_training}

Next, we investigate if NLI datasets provide a good signal during training.
To this end, we pre-train Llama-3 architecture-based 8B and 70B models from scratch for 2T tokens. The details of the pre-training setup are provided in \cref{appx:experiments}.

\begin{figure*}[t]
    \begin{subfigure}[b]{0.20\textwidth}
    \centering
    \includegraphics[height=3.2cm]{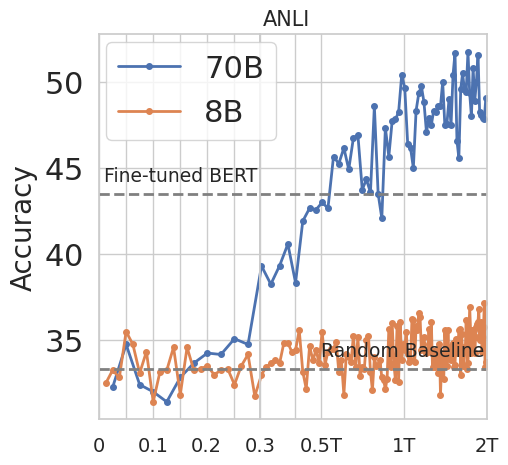}
    \caption{ANLI}
    \end{subfigure}
    \label{fig:anli_int}
    \begin{subfigure}[b]{0.19\textwidth}
    \centering
    \includegraphics[height=3.2cm, trim=11mm 0 0 0, clip]{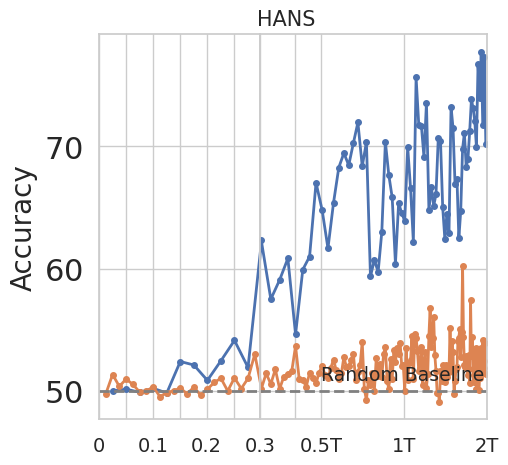}
    \caption{HANS}
    \label{fig:hans_int}
    \end{subfigure}
    \begin{subfigure}[b]{0.19\textwidth}
    \centering
    \includegraphics[height=3.2cm, trim=11mm 0 0 0, clip]{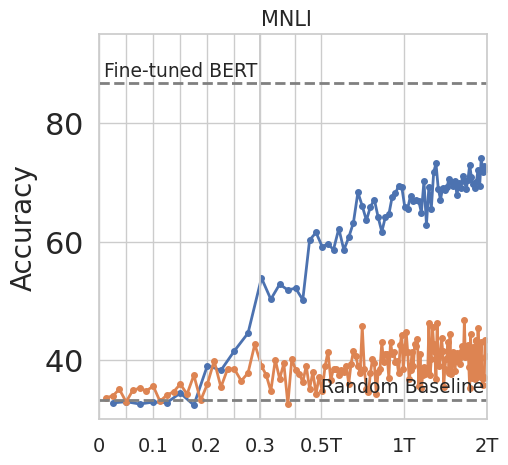}
    \caption{MNLI}
    \label{fig:mnli_int}
    \end{subfigure}
    \begin{subfigure}[b]{0.19\textwidth}
    \centering
    \includegraphics[height=3.2cm, trim=11mm 0 0 0, clip]{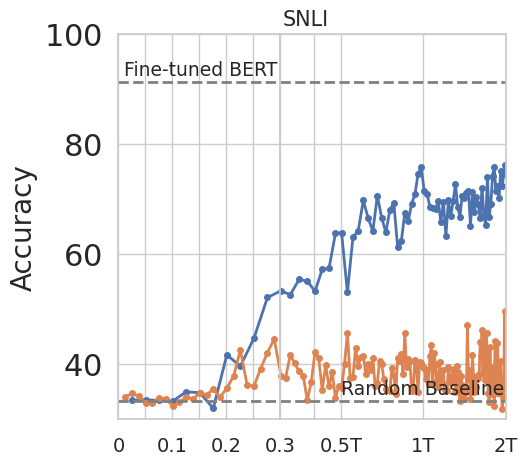}
    \caption{SNLI}
    \label{fig:snli_int}
    \end{subfigure}
    \begin{subfigure}[b]{0.19\textwidth}
    \centering
    \includegraphics[height=3.2cm, trim=11mm 0 0 0, clip]{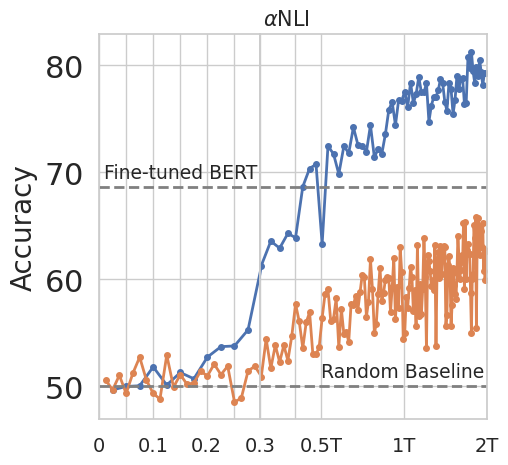}
    \caption{$\alpha$NLI}
    \label{fig:alphanli_int}
    \end{subfigure}
    \caption{\textbf{Performance during pre-training}. 
    We show how accuracy for the five benchmarks develops during pre-training for two Llama-3 style models.}\label{fig:performance_training}
\end{figure*}

\paragraph{Training curves}
In \cref{fig:performance_training}, we show how four-shot performance develops during training.
We see that for most benchmarks, the 8B and 70B model quickly start to diverge.
The 70B model starts improving after 250B tokens for ANLI and $\alpha$NLI; it crosses fine-tuned BERT performance after 500B tokens.
On the other hand, the development of performance for the 8B model is slow, not exceeding chance accuracy for HANS, MNLI, and SNLI.
From the final model performance of 8B depicted in \cref{fig:shot_performance}, we can conclude that the 8B model improves fairly late in pre-training.
We did not have the budget for a full pre-training run, but longer training seems to help for NLI tasks, further supporting the claim that NLI models can provide a useful signal.

\begin{table}
\centering
\resizebox{0.45\textwidth}{!}{
\begin{tabular}{lcccc}
    & \multicolumn{2}{c}{8B} & \multicolumn{2}{c}{70B} \\
    & $mon_{Acc}$  & $mon_{NLL}$ & $mon_{Acc}$ & $mon_{NLL}$ \\
    \toprule
    $\alpha$NLI & 0.62 & 0.62 & 0.79 & 0.79 \\
    \midrule
    ANLI & - & - & 0.67 & 0.47 \\
    \midrule
    HANS & 0.32 & 0.46 & 0.57 & 0.63 \\
    \midrule
    MNLI & 0.34 & 0.51 & 0.77 & 0.80 \\
    \midrule
    SNLI & 0.05 & 0.38 & 0.64 & 0.65 \\
    \bottomrule
    \end{tabular}
}
    \caption{\textbf{Monotonicity values during the course of training.} We report monotonicity both for accuracy ($mon_{Acc}$) and negative log likelihood of the correct answer ($mon_{NLL}$).}
\label{tab:monotonicity}
\end{table}

\paragraph{Utility for ablations}
A requirement for a benchmark to provide a useful signal during training is that it develops relatively monotonically during training.
The plots in \cref{fig:performance_training} suggest that this is not the case for most of the benchmarks for the 8B model.
Following \citet{variancepaper}, we quantify the benchmarks' monotonicity -- defined as the rank correlation (Kendall Tau) between a monotonically increasing array and the array containing the benchmark scores during training -- on both a discrete (accuracy) and continuous metric (NLL). 
In \cref{tab:monotonicity}, we can see that, despite the benchmarks' moderate sizes, monotonicity is low for accuracy as well as NLL, suggesting that the benchmark may not be suitable for monitoring performance on closely-spaced checkpoints at this scale. 

\subsection{Contamination analysis}\label{subsec:contamination}

Now that we have concluded that NLI benchmarks provide a signal both for final models and during pre-training, we analyse the extent to which scores may be driven by evaluation data contamination.
Using the methodology of \citet{singh2024evaluationdatacontaminationllms} and \citet{dubey2024llama}, we analyse contamination by assigning a score to each evaluation sample which represents the percentage of tokens in the sample that is part of an 8-gram occurring in the pre-training datamix.
Following \citet{singh2024evaluationdatacontaminationllms}, we select contamination thresholds using \texttt{ConTAM}, considering for each threshold the estimated performance gain (EPG), defined as the difference in performance on the full evaluation set and the `uncontaminated' subset.\footnote{A threshold of 0 implies that all examples with non-zero contamination scores to be marked as contaminated, as we increase the threshold, examples are moved to the `uncontaminated' subset.}
Since EPG is model-dependent, we use the 8B and 70B models we trained from scratch to observe the effect of contamination. 

\begin{figure*}[t]
    \begin{subfigure}[b]{0.20\textwidth}
    \centering
    \includegraphics[height=2.6cm]{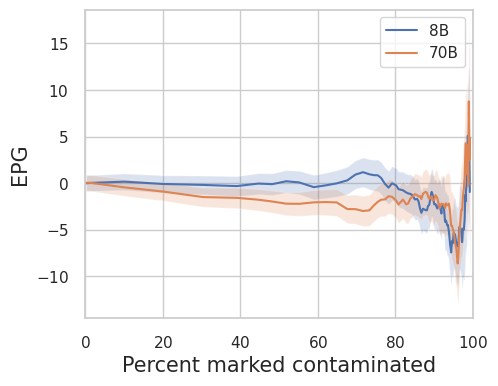}
    \caption{ANLI}
    \end{subfigure}
    \label{fig:cont_anli}
    \begin{subfigure}[b]{0.19\textwidth}
    \centering
    \includegraphics[height=2.6cm]{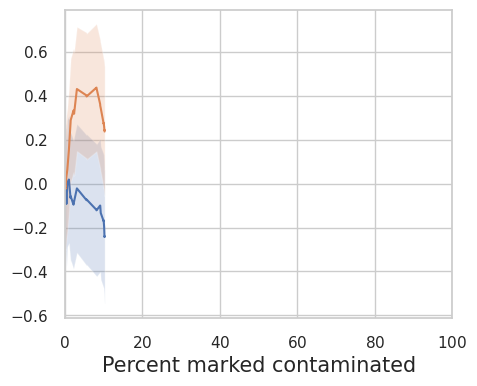}
    \caption{HANS}
    \label{fig:cont_hans}
    \end{subfigure}
    \begin{subfigure}[b]{0.19\textwidth}
    \centering
    \includegraphics[height=2.6cm]{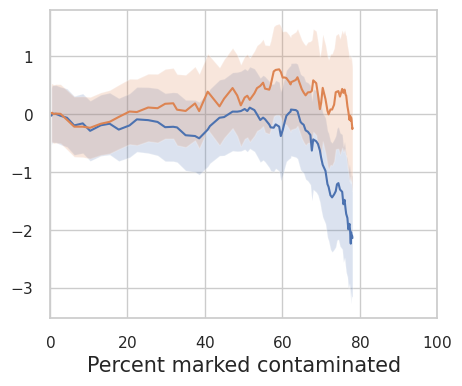}
    \caption{MNLI}
    \label{fig:cont_mnli}
    \end{subfigure}
    \begin{subfigure}[b]{0.19\textwidth}
    \centering
    \includegraphics[height=2.6cm]{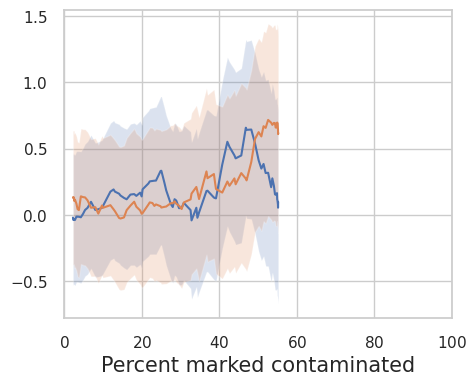}
    \caption{SNLI}
    \label{fig:cont_snli}
    \end{subfigure}
    \begin{subfigure}[b]{0.19\textwidth}
    \centering
    \includegraphics[height=2.6cm]{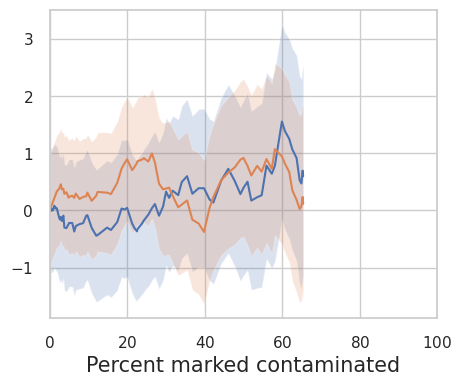}
    \caption{$\alpha$NLI}
    \label{fig:cont_alphanli}
    \end{subfigure}
    \caption{\textbf{Contamination results.} We show the EPG vs the percent of the evaluation dataset marked as contaminated according to different thresholds.}\label{fig:contamination}
\end{figure*}

In \cref{fig:contamination}, we plot the EPG as a function of the percentage of the evaluation data that is marked contaminated according to a given contamination threshold.
We can see that there is virtually no score inflation as a consequence of contamination for the 8 and 70B model that we considered.
For HANS, this is because little to no contaminated samples are found through 8-gram overlap.
For MNLI, SNLI, and $\alpha$NLI, instead, large parts of the dataset are marked contaminated at low thresholds, but there is no performance impact.
We suspect that this is because the \textit{premises} of the samples are sourced from publicly available datasets, and their presence per se is thus not indicitive of contamination.
Likely for similar reasons, almost all of the ANLI samples contain at least one overlapping 8-gram, resulting in very high percentages of detected contamination at low thresholds.
At the lowest threshold, also the EPG shoots up.
However, the erratic behaviour as examples are gradually added to the clean set suggests that the EPG observed at those threshold is likely an artefact of the small size of the clean partition, and this result is likely not indicative of true performance gain.
Thus, we believe that contamination does not play a participatory role in high performances for NLI benchmarks. 

\subsection{Dataset saturation}\label{subsec:saturation}
Having confirmed that the NLI benchmarks under scrutiny provide discrimination between LLMs of different sizes and are not affected much by contamination, we now turn to the question of saturation: the results show that the benchmarks would have been useful so far, but how about the future?
As already pointed out before, the benchmark that still has the clearest room for improvement is adversarial benchmark ANLI, with performances not exceeding 70\% even for the largest models.
For the other benchmarks, performances are substantially higher, and it is unclear to what extent the benchmarks may suffer saturation.

\begin{table*}[h!]
\centering
\small
\resizebox{\textwidth}{!}{
\begin{tabular}{p{10.1cm}|c|c|c}
\textbf{Example} & \textbf{Label distribution} & \textbf{Gold} & \textbf{Prediction} \\
\midrule
    \makecell[l]{
    \textbf{Premise}: of course you could annex Cuba but they wouldn't like that a bit\\
    \textbf{Hypothesis}: Annexing Cuba is a great idea.
    } & e: 0, n: 31, \textbf{c: 69} & c & n \\
\midrule
\makecell[l]{
    \textbf{Premise}: I thought working on Liddy's campaign would be better than working\\ on Bob's. \\
    \textbf{Hypothesis}: I thought I would like working on Liddy's campaign the best.
    } & \textbf{e: 66}, n: 32, c: 2 & n & e \\
\midrule
\makecell[l]{
\textbf{Premise}: Sorry but that's how it is.\\
\textbf{Hypothesis}: This is how things are and there are no apologies about it.
    }& \textbf{e: 48}, c: 40, n: 12 & c & c \\
\end{tabular}
    }
    \caption{\textbf{Examples from ChaosNLI with different label distributions.} We show different samples from ChaosNLI with different annotator distributions. The first column shows the example; the second the distribution of human labels across the 100 collected annotations (the majority label is marked bold), the third column indicates the label that the example had in the original dataset and the fourth column the Llama 3.1 405B prediction.}
\label{tab:sample_analysis}
\end{table*}

\paragraph{Error types} To address this question, we first conduct a manual error analysis on the examples that the largest models assign an incorrect label.
Specifically, we focus on MNLI, one of the datasets with the highest scores and analyse 40 predictions of Llama 3.1 405B model.
In \cref{tab:sample_analysis}, we show examples, along with distribution of the 100 human annotations for that sample from the previously mentioned dataset ChaosNLI.
For all the examples we investigated, we found that there was at least some degree of interpretation or ambiguity in assigning a label to the premise-hypothesis pair.
For example, in the first row of \cref{tab:sample_analysis}, whether the correct label is contradiction or negation would depend on a not-specified sentiment towards Cuba.
In fact, 31 out of 100 human annotators would agree with the model's judgement, despite it being labeled as ``incorrect''.
Also in the second row of \cref{tab:sample_analysis}, we see an example where human opinions diverge on what the correct label is.
In this case, the model's prediction matches the majority label found by \citet{nie-etal-2020-learn}, but the sample is marked incorrect because the MNLI label is different.
It is worth pointing out, that the same is sometimes true for examples that are marked \emph{correct}.
Consider, for instance, the last row in \cref{tab:sample_analysis}, where the model prediction matches the gold label in the MNLI dataset, but not the majority label collected by ChaosNLI.
In sum, it appears that for the best model, most of the `mistakes' in MNLI are cases in which humans may not agree on the correct label.

\begin{table*}
    \centering
    \begin{tabular}{llcccccc}
        & & \multicolumn{2}{c}{\textbf{$\alpha$NLI}} & \multicolumn{2}{c}{\textbf{MNLI}} & \multicolumn{2}{c}{\textbf{SNLI}} \\
        \multicolumn{2}{c}{\textbf{Model}} & \textit{Og.} & \textit{Maj.} & \textit{Og.} & \textit{Maj.} & \textit{Og.} & \textit{Maj.}\\
        \toprule
        Llama-3.1 & 8B & 77.55 & \textbf{78.00} & 49.47 & \textbf{50.97} & \textbf{55.28} & 55.48 \\
        Llama-3.1 & 70B & 86.36 & \textbf{87.21} & 57.66 & \textbf{67.54} & \textbf{60.44} & 58.52 \\
        Llama-3.1 & 405B & 85.51 & \textbf{86.10} & 64.04 & \textbf{69.67} & 64.60 & \textbf{67.31} \\
        \midrule
        Mistral & 7B & 74.41 & \textbf{75.78} & 49.97 & \textbf{53.22} & \textbf{49.47} & 48.15 \\
        Mixtral & 8x7B & \textbf{82.44} & 81.59 & \textbf{54.03} & 51.53 & 63.14 & \textbf{64.27} \\
        Mixtral & 8x22B & \textbf{84.14} & 83.68 & 60.23 & \textbf{67.04} & 64.86 & \textbf{67.83} \\
        \midrule
        \multicolumn{2}{c}{\emph{Average}} & 76.01 & 76.21 & 50.86 & 55.93 & 54.90 & 54.35 \\
    \end{tabular}
    \caption{\textbf{Majority accuracy.} For each model and dataset, we show the accuracy on the ChaosNLI subsets of $\alpha$NLI, MNLI, and SNLI. The original accuracy (Og.) represents the accuracy as computed according to the original labels of the respective datasets; the majority accuracy (Maj.) expresses accuracy according to the majority label.}
\label{tab:chaos_acc}
\end{table*}

\paragraph{Majority accuracy} 
Next, we study this phenomenon more quantitatively, again utilising the ChaosNLI dataset which contains 100 human annotations for over 1500 samples each for MNLI, SNLI, and $\alpha$NLI.
First, we consider how model accuracies change when we replace the original labels with the majority label of the ChaosNLI dataset.
This alters 32\%, 25\%, and 11\% of the labels of MNLI, SNLI, and $\alpha$NLI, respectively.
In Table \ref{tab:chaos_acc}, we can see that the results differ per model and benchmark.
The largest effect is observed for MNLI, where for some models, there's an increase of more than 10\% in performance and the average accuracy across models is more than five points higher on the `corrected' datasets.
For the other two benchmarks, the results are more mixed, with little to no difference on average.
Only for the largest Llama-3.1 model (405B), the majority accuracy is systematically higher than the original accuracy, suggesting it may have honed in more on the majority label.
Interestingly, the MNLI and SNLI subsets of ChaosNLI appear substantially more difficult than the average dataset; even the Llama-3.1 405B model stays below 70\% for both these subsets, suggesting that there is room for improvement.

\begin{figure*}
    \centering
    \begin{subfigure}[b]{0.3\textwidth}
        \begin{subfigure}[b]{0.48\textwidth}
        \includegraphics[height=2.2cm, trim=10 30 0 0, clip]{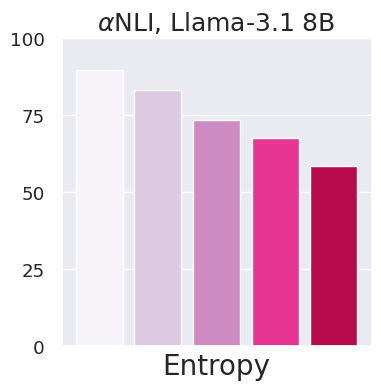}
        \end{subfigure}
        \begin{subfigure}[b]{0.35\textwidth}
            \includegraphics[height=2.2cm, trim=10 30 0 0, clip]{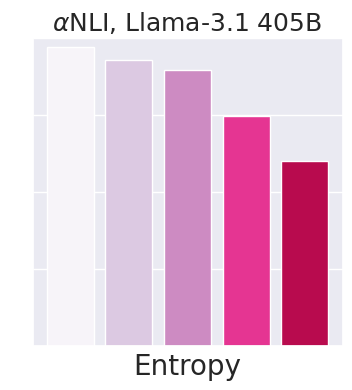}
        \end{subfigure}\\
        \begin{subfigure}[b]{0.48\textwidth}
            \includegraphics[height=2.2cm, trim=10 30 0 0, clip]{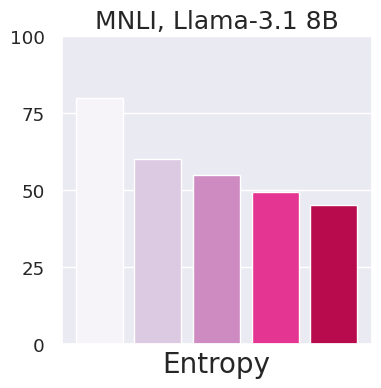}
        \end{subfigure}
        \begin{subfigure}[b]{0.35\textwidth}
            \includegraphics[height=2.2cm, trim=10 30 0 0, clip]{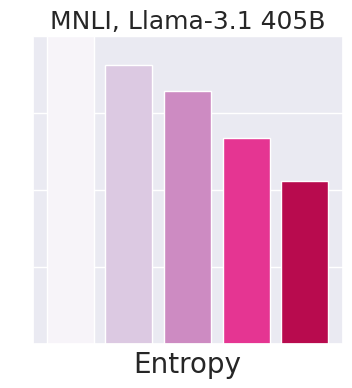}
        \end{subfigure}\\
        \begin{subfigure}[b]{0.48\textwidth}
            \includegraphics[height=2.4cm, trim=10 0 0 0, clip]{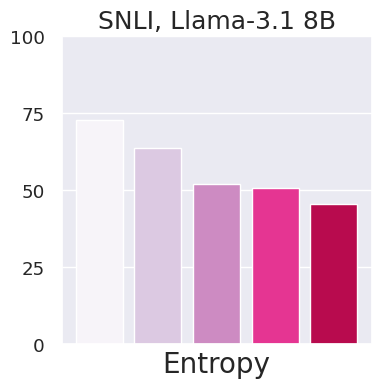}
        \end{subfigure}
        \begin{subfigure}[b]{0.35\textwidth}
            \includegraphics[height=2.4cm, trim=10 0 0 0, clip]{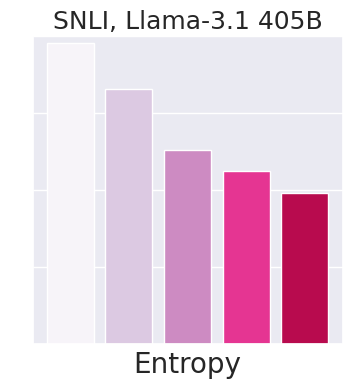}
        \end{subfigure}
        \caption{Accuracy vs Entropy}
         \label{fig:entropy_accuracy}
    \end{subfigure}
    \hspace{5mm}
    \begin{subfigure}[b]{0.65\textwidth}
    \includegraphics[width=\linewidth]{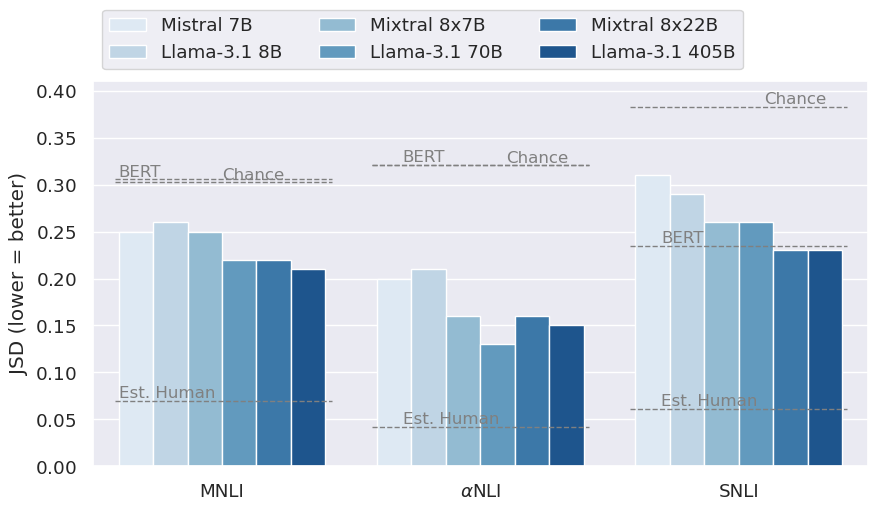}
    \caption{Final model JSD}
    \label{fig:jsd_all_benchmarks}
    \end{subfigure}
    \caption{\textbf{Accuracy vs entropy and final model JSDs} 
    a) Accuracy vs entropy for Llama8B and Llama 405B. 
     We show how the accuracy of Llama 8B and Llama 405B changes as the entropy of the human label distributions increases. Accuracy-vs-entropy plots for all other models can be found in \cref{fig:entropy_accuracy_all}.
    b) Final model JSDs for each of the benchmarks in ChaosNLI. 
    JSDs are substantially lower than chance and BERT JSDs, but substantially higher than JSDs between humans.
    }
\end{figure*}

\paragraph{Accuracy versus entropy}
Next, we consider the distribution of accuracy with entropy for the three benchmarks in Figure \ref{fig:entropy_accuracy}. 
We only show the smallest (8B) and the largest (405B) model in the Llama series across the three benchmarks. 
We observe that for the largest model, there is a clear decreasing trend in accuracy with increasing entropy for all three benchmarks, whereas for the 8B model, the trend is not as prominent, and some entropy bins have similar accuracies. 
In \cref{fig:entropy_accuracy}, we can furthermore see that all models perform `better' on samples where the entropy of the labels is low.
For the larger models, this effect is larger on samples where humans have high disagreement (and the majority label is thus in a way more representative of the average human judgements), their accuracies are often near maximal, and they drop as human judgements become more dispersed.
For the entropy vs accuracy distributions for other models, we refer the reader to Appendix \ref{app:entropy_vs_accuracy}.

\paragraph{Implications for saturation}
What the implication of these results is for the question of saturation of these benchmarks is, like the samples of the benchmark themselves, open to interpretation.
On the one hand, it appears that for several cases, the model predictions do not align with the human majority label, suggesting that the performance may still improve as models continue training.
On the other, the results put into question whether strong alignment with the majority label is in fact what we should strive for: if humans have low agreement on their judgements, should the desired behaviour of a model be to side with the majority?
In the next section, we approach this question by considering the distribution of model outputs, rather than only the highest probability label.

%
%

\begin{figure*}[t]
    \centering
    \begin{subfigure}[b]{0.32\textwidth}
        \includegraphics[height=4.5cm]{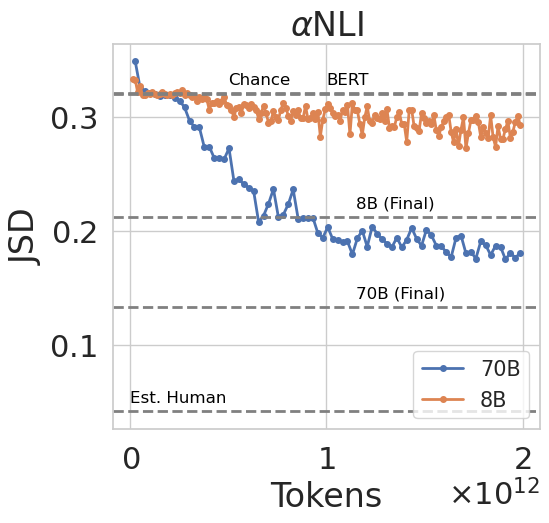}
    \end{subfigure}
    \begin{subfigure}[b]{0.32\textwidth}
        \includegraphics[height=4.5cm, trim=11mm 0 0 0, clip]{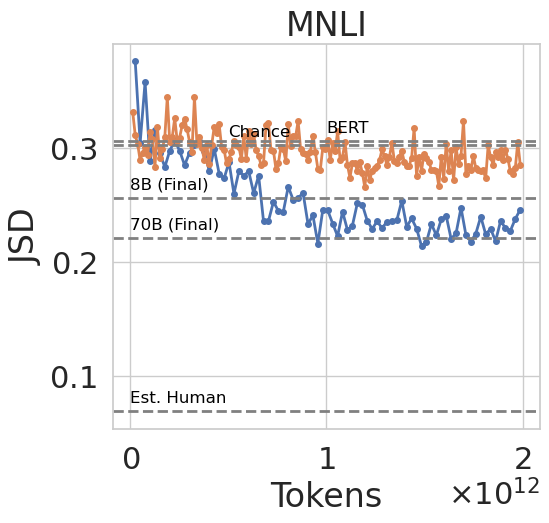}
    \end{subfigure}
    \begin{subfigure}[b]{0.32\textwidth}
        \includegraphics[height=4.5cm, trim=11mm 0 0 0, clip]{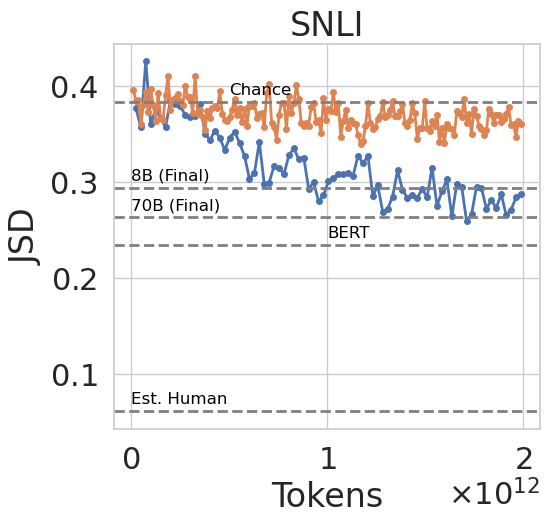}
    \end{subfigure}
    \caption{\textbf{Development of JSD during training.} We show how the JSD of our trained-from-scratch 8B and 70B model develops during training.}\label{fig:jsd_training}
\end{figure*}

\subsection{Model versus human distributions}\label{subsec:chaosnli_dist}
The ChaosNLI dataset does not only help us estimate the adequacy of the original label sets of the respective datasets, but also reveals model behaviour in scenarios without a single correct answer.
In a time where LLMs are deployed for many users with potentially different preferences, this question has become very practically relevant.
To do so, we consider how the probability distribution of the models over the three possible labels \textit{Entailment}, \textit{Neutral}, and \textit{Contradiction}) compares with the label distributions observed in the human annotations.
Following \citet{nie-etal-2020-learn}, we consider Jensen-Shannon Divergence \citep[JSD][]{menendez1997jensen} to measure the distance between the two distributions. 
%
%
Contrary to KL divergence, JSD is symmetric and bound between 0 and 1, making it more interpretable for our specific case.

In \cref{fig:jsd_all_benchmarks}, we can see that, for MNLI, all models have probability distributions more similar to the human distributions than chance, but substantially different than humans have among each other.\footnote{The human estimate was computed by \citet{nie-etal-2020-learn}, on an independent sample of human labels on the same data.}
This pattern is constant across datasets.
Interestingly, larger models have lower divergence, contradicting the finding of \citet{nie-etal-2020-learn} that better or larger models do not have more similar distributions.
Yet, the effect of scale is further confirmed in \cref{fig:jsd_training}: though even trained-out models are far away from a low JSD, the JSD of the model softmax distributions and the human label distributions decreases steadily during training.
This is an exciting finding, since it suggests that measuring similarity with human distributions could be an interesting venue to explore during training.

\section{Conclusion}

In this work, we revisit natural language inference (NLI) benchmarks and investigate if they may still play a role in LLM evaluation, both during and after pre-training.
We consider five different NLI benchmarks -- $\alpha$NLI, ANLI, HANS, and MNLI -- and evaluate them across six different models of two model families: Llama 3.1 8B, 70B, and 405B, and Mistral 7B, 8x7B, and 8x22B.
Furthermore, we consider how the benchmark behave during the training of two Llama-style 8B and 70B models.
We find that, with the exception of $\alpha$NLI, all benchmarks are able to discriminate between models of different qualities, and in particular ANLI is challenging even for the largest models.
Furthermore, we find next to no effect of contamination for the benchmarks.
Considering the benchmark ChaosNLI \citep{nie-etal-2020-learn}, containing 100 human annotations for over 4500 samples of three of the benchmarks we consider, we also find that the differences between human label distributions and model label distributions -- as measured with Jensen Shannon Divergence (JSD) -- has decreased for the new generation of models.
However, they are still substantially higher than the distributional difference between two populations of humans. 
Interestingly, contrary to the findings of \citet{nie-etal-2020-learn}, we observe a clear effect of scale.
JSD shows a steady decrease during training, and larger models have lower JSD than smaller models, making it a potentially interesting quality to consider for model development.

\bibliography{anthology,custom}

\appendix

\section{Benchmarks}
\label{app:benchmarks}

Below, we provide a more elaborate description of (construction of) the benchmarks we consider in our work.
A summary of their statistics can be found in \cref{tab:dataset}.

\begin{table}
\centering
\begin{tabular}{c|c|c}
    Benchmark & \# Samples & \# labels \\
    \hline
    ANLI & 1200 & 3 \\
    HANS & 30000 & 2 \\
    MNLI & 9815 & 3 \\
    SNLI & 9842 & 3 \\
    $\alpha$NLI & 1532 & 2 \\
\end{tabular}
\caption{Dataset details}
\label{tab:dataset}
\end{table}

\subsection{SNLI}
Introduced by \citet{bowman-etal-2015-large}, the Stanford Natural Language Inference (SNLI) dataset was one of the first large-scale NLI dataset for NLP evaluation.
The dataset comprises of 570K human-authored English sentence pairs, sourced by asking Amazon Mechanical Turk workers to supply hypotheses for the three labels available in the dataset -- entailment, neutral and contradiction -- given a premise comprised by an image caption drawn from a pre-existing corpus.
For 57K of the resulting samples were then labeled by four additional annotators.
In this work, we consider the 10K development set of the corpus.
Like the original paper, we exclude samples with no gold label because there was no label that the annotators agreed on.

\subsection{MNLI}
The Multi-Genre NLI (MNLI) corpus \citep{williams-etal-2018-broad} was introduced as an alternative to SNLI that captures more genres and more challenging examples, representing both written and spoken speech in a range of different styles, degrees, formalities, and topics.
The data collection procedure of the corpus is similar to the SNLI procedure both in terms of sourcing and validation.
Unlike SNLI, the MNLI premise sentences are derived from nine different sources, aiming to represent the full range of American English rather than a single image captioning corpus.
As for SNLI, we consider the validation corpus of the dataset and exclude samples that have no gold label.

\subsection{HANS}
Deviating from previous datasets, the adversarial dataset Heuristic Analysis for NLI Systems \citep[HANS,][]{mccoy-etal-2019-right} is not crowd-sourced, but synthetically generated using templates.
Specifically, the templates are designed to generate examples that can not be solved through heuristics such as lexical, subsequence, or constituent overlap.
At the time of proposal, none of the SOTA models were able to solve such examples.

\subsection{ANLI}
The second adversarial dataset we consider is Adversarial NLI \citep[or ANLI,][]{nie-etal-2020-adversarial}.
The dataset, created with the primary aim to make SOTA models fail, is sourced iteratively in a human-in-the-loop setup.
Given a premise and a target label, annotators are asked to propose hypotheses that may fool models.
The produced samples are then tested on a model, and the examples that do indeed receive an incorrect label are re-validated by one or more human validators.
The dataset consists of three sets of increasingly challenging examples, where in each round more powerful models are considered that are trained on examples from the previous round.
The third round furthermore contains a set of more diverse premises.
For our experiments, we are using the dev set of round 3, the most challenging set of the benchmark.

\subsection{$\alpha$NLI}
Differing in setup from the previously described benchmarks, $\alpha$NLI or abductive NLI \citep{bhagavatula2020abductive} focuses on \emph{abductive reasoning} -- which they describe as the inference of the most plausible explanation for an incomplete observation.
The samples in $\alpha$NLI consist of a pair of observations at two consecutive times, and a plausible hypothesis that explains tho two observations, and an implausible hypothesis that does not (or to a lesser extent).
The task is to select the most plausible hypothesis.
To construct the data \citet{bhagavatula2020abductive} first draw observation pairs from a stories dataset and then ask crowd-sources to generate plausible and implausible hypotheses. 
For each observation pair, multiple plausible and implausible hypotheses are crowd-sourced, and adversarial filtering is applied to retain one challenging pair of hypotheses.
We use the development set of the corpus for our experiments.

\begin{figure*}[t]
    \centering
    \begin{subfigure}[b]{0.22\textwidth}
        \includegraphics[height=3.4cm]{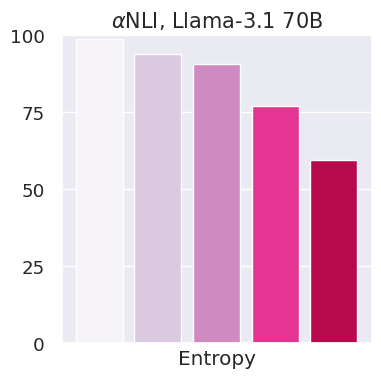}
    \end{subfigure}
    \begin{subfigure}[b]{0.22\textwidth}
        \includegraphics[height=3.4cm]{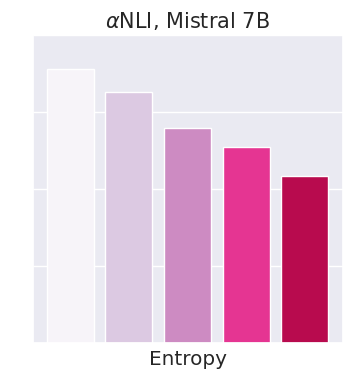}
    \end{subfigure}
    \begin{subfigure}[b]{0.22\textwidth}
        \includegraphics[height=3.4cm]{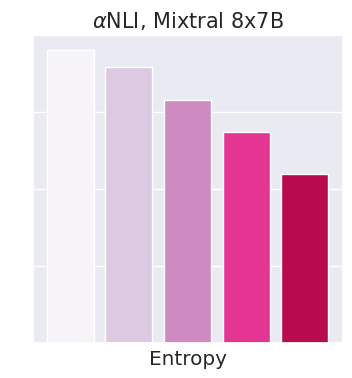}
    \end{subfigure}
    \begin{subfigure}[b]{0.22\textwidth}
        \includegraphics[height=3.4cm]{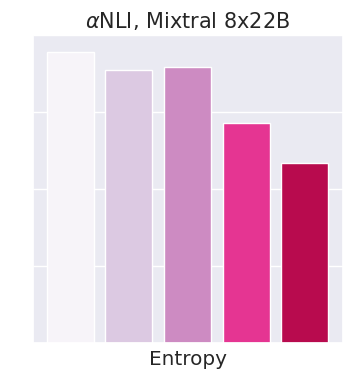}
    \end{subfigure}\\
    \begin{subfigure}[b]{0.22\textwidth}
        \includegraphics[height=3.4cm]{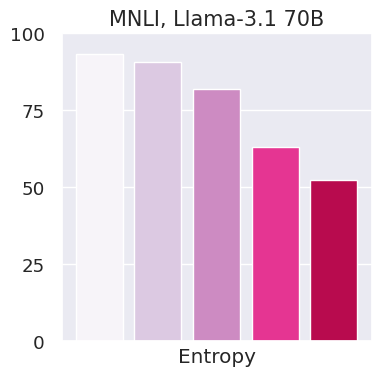}
    \end{subfigure}
    \begin{subfigure}[b]{0.22\textwidth}
        \includegraphics[height=3.4cm]{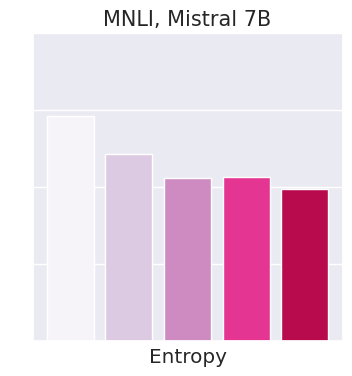}
    \end{subfigure}
    \begin{subfigure}[b]{0.22\textwidth}
        \includegraphics[height=3.4cm]{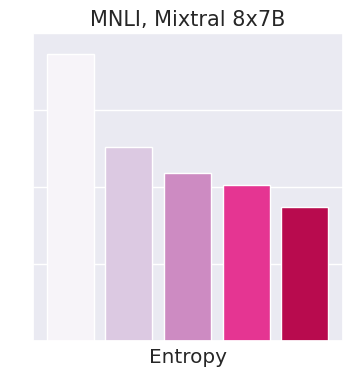}
    \end{subfigure}
    \begin{subfigure}[b]{0.22\textwidth}
        \includegraphics[height=3.4cm]{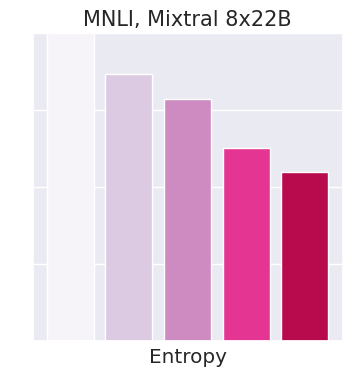}
    \end{subfigure}\\
    \begin{subfigure}[b]{0.22\textwidth}
        \includegraphics[height=3.4cm]{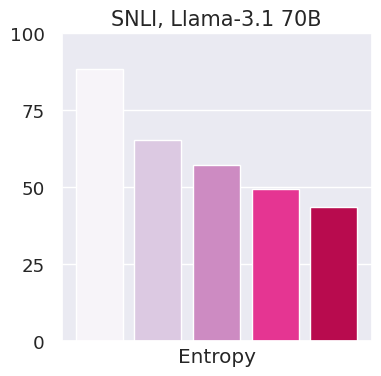}
    \end{subfigure}
    \begin{subfigure}[b]{0.22\textwidth}
        \includegraphics[height=3.4cm]{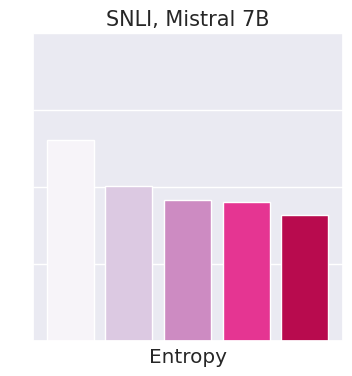}
    \end{subfigure}
    \begin{subfigure}[b]{0.22\textwidth}
        \includegraphics[height=3.4cm]{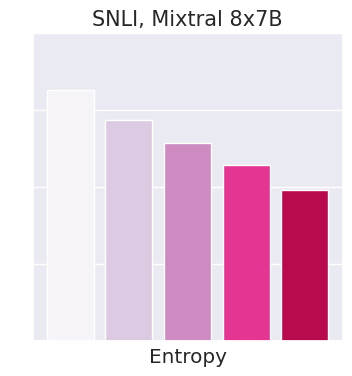}
    \end{subfigure}
    \begin{subfigure}[b]{0.22\textwidth}
        \includegraphics[height=3.4cm]{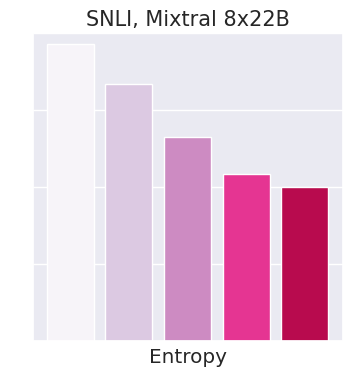}
    \end{subfigure}
    \caption{\textbf{Accuracy versus entropy.} We show how the accuracy of Llama-3.1 70B and the Mistral series of models changes as the entropy of the human label distributions increases.}
    \label{fig:entropy_accuracy_all}
\end{figure*}

\section{Entropy vs accuracy plots}\label{app:entropy_vs_accuracy}

In addition to \cref{fig:entropy_accuracy} highlighting the results on the Llama-3.1 8B and 405B models, we also show the accuracy vs entropy plots of all other models we use for our analysis: Llama-3.1 70B, Mistral 7B, Mixtral 8x7B, and Mixtral 8x22B in \cref{fig:entropy_accuracy_all}. The Mistral family of models also show a similar trend where larger models have lower accuracies for higher entropy buckets.

\section{Prompt templates}

The prompt templates used for each task are presented in \cref{tab:prompt_template}.

\begin{table*}[t]
    \centering
    \small
    \begin{tabular}{lp{8cm}}
        \textbf{Benchmark} & \textbf{Prompt Template} \\
        \midrule
        MNLI, SNLI, ANLI & \begin{verbatim}
{% for x in few_shot -%}
Premise: {{ x["premise"] }}
Hypothesis: {{ x["hypothesis"] }}
A. Entailment
B. Neutral
C. Contradiction
Answer: {{ x["answer"] }}

{% endfor -%}
Premise: {{ premise }}
Hypothesis: {{ hypothesis }}
A. Entailment
B. Neutral
C. Contradiction
Answer: {{ choice_text }}
\end{verbatim} \\
\midrule
AbductiveNLI & \begin{verbatim}
{% for x in few_shot -%}
Observation 1: {{ x["obs1"] }}
Observation 2: {{ x["obs2"] }}
A. {{ x["choices"]["A"] }}
B. {{ x["choices"]["B"] }}
Answer: {{ x["answer"] }}

{% endfor -%}
Observation 1: {{ obs1 }}
Observation 2: {{ obs2 }}
A. {{ choices["A"] }}
B. {{ choices["B"] }}
Answer: {{ choice_text }}
\end{verbatim} \\
\midrule
HansNLI & \begin{verbatim}
{% for x in few_shot -%}
Premise: {{ x["premise"] }}.
Hypothesis: {{ x["hypothesis"] }}.
A. Entailment
B. Non-Entailment
Answer: {{ x["answer"] }}

{% endfor -%}
Premise: {{ premise }}.
Hypothesis: {{ hypothesis }}.
A. Entailment
B. Non-Entailment
Answer: {{ choice_text }}
\end{verbatim} \\
\end{tabular}
\caption{Prompt Templates for each task}
\label{tab:prompt_template}
\end{table*}

\section{Experimental Details}
\label{appx:experiments}

For the 8B and 70B models we pre-train from scratch, we use our custom pre-training datamix, a mix of data available from publicly available sources including web data, code, and reasoning datasets. 
For pre-training hyperparameters, we use similar settings as reported in \citet{dubey2024llama}. 
We use a batch size of 4M tokens and pre-train the models for 500,000 steps, resulting in a total of 2T token training. 
We use 512 GPUs for a single pre-training run of both models.

For running the evaluations, we use 16 GPUs for each model comprising all five NLI benchmarks and different shot settings in a single job. 
A single evaluation job takes on average takes around 55 mins for the five benchmarks.

\end{document}